\documentclass[runningheads]{llncs}

\usepackage{eccv}

\usepackage{eccvabbrv}

\usepackage{graphicx}
\usepackage{booktabs}

\usepackage{amsmath}
\usepackage{amssymb}
\usepackage{xcolor}
\usepackage{enumitem}
\usepackage{graphicx}
\usepackage{wrapfig}
\usepackage{graphicx}
\usepackage{url}
\usepackage{adjustbox}
\usepackage{multirow}
\usepackage{boldline}

\usepackage[accsupp]{axessibility}  %

\usepackage{hyperref}

\usepackage{orcidlink}

\begin{document}

\title{An Investigation on The Position Encoding  in \\ Vision-Based Dynamics Prediction} 

\titlerunning{Position Encoding in Dynamics Prediction}

\author{Jiageng Zhu*\inst{1,2} \orcidlink{0009-0002-0162-6534} \and
Hanchen Xie*\inst{1,3} \orcidlink{0009-0004-4474-4877} \and
Jiazhi Li\inst{1,3} \orcidlink{0000-0003-3938-7989} \and Mahyar Khayatkhoei\inst{1} \orcidlink{0000-0002-7326-861X} \and Wael AbdAlmageed\inst{4} \orcidlink{0000-0002-8320-8530} }

\authorrunning{J. Zhu et al.}

\institute{USC Information Sciences Institute \and
USC Ming Hsieh Department of Electrical and Computer Engineering \and
USC Thomas Lord Department of Computer Science \and Clemson University Holcombe Department of Electrical and Computer Engineering }

\maketitle

\let\thefootnote\relax\footnotetext{*: Equal Contributions}

\begin{abstract}

Despite the success of vision-based dynamics prediction models, which predict object states by utilizing RGB images and simple object descriptions, they were challenged by environment misalignments. Although the literature has demonstrated that unifying visual domains with both environment context and object abstract, such as semantic segmentation and bounding boxes, can effectively mitigate the visual domain misalignment challenge, discussions were focused on the abstract of environment context, and the insight of using bounding box as the object abstract is under-explored. Furthermore, we notice that, as empirical results shown in the literature, even when the visual appearance of objects is removed, object bounding boxes alone, instead of being directly fed into the network, can indirectly provide sufficient position information via the Region of Interest Pooling operation for dynamics prediction. However, previous literature overlooked discussions regarding how such position information is implicitly encoded in the dynamics prediction model. Thus, in this paper, we provide detailed studies to investigate the process and necessary conditions for encoding position information via using the bounding box as the object abstract into output features. Furthermore, we study the limitation of solely using object abstracts, such that the dynamics prediction performance will be jeopardized when the environment context varies.

\end{abstract}

\begin{figure*}[]
\centering
     \includegraphics[width=1.0\textwidth]{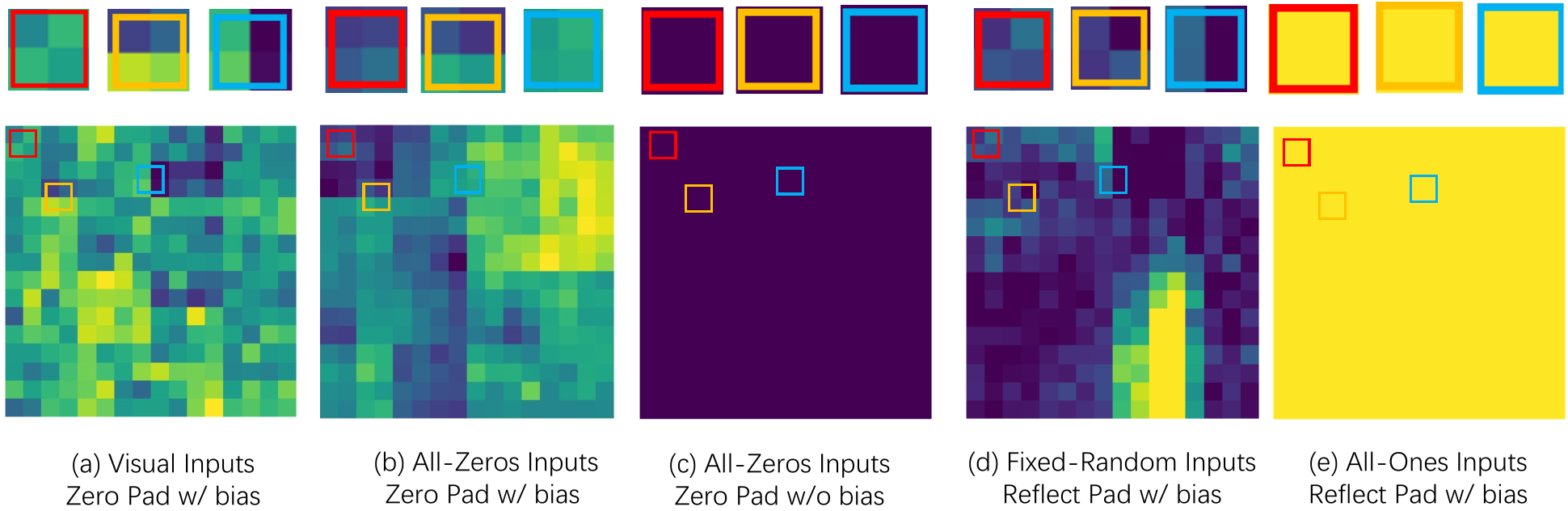}
    \caption{Illustration of output features of Hourglass module with various inputs and padding settings, and depiction of the 
    process of position information encoding utilizing bounding-box and RoI Pooling operation. The position information can be encoded when there is inconsistency in the distribution of output features of the Hourglass module, where different values across fragmented object state features are incorporated to encode position information.
    Such inconsistency can be brought up by either proper padding settings or discrepancies existing in original inputs.} %
    \label{fig:illustration}
\end{figure*}

\section{Introduction}\label{sec:introduction}

Dynamics prediction \cite{compo_obj_base_phys, reasoing_phys_interaction, interaction_learn_phys, rpcin, bib:sim-blen-datasets}, which aims at predicting the state of the object of interest in the future by referencing previous states, has drawn increasing attention. Physics-state-based models \cite{interaction_learn_phys, compo_obj_base_phys} take well-defined physics parameters, such as position, mass, and velocity, as inputs and derive the future state via pre-defined physics models \cite{end_to_end_diff_phys, ullman2014learning, dynamic_reason_diff_phys} or deep neural networks (DNNs) \cite{compo_obj_base_phys, visual_de_animation, interaction_learn_phys}. However, since the visual information is completely absent from dynamics prediction, such steam of methods is limited and challenged when being applied to real-world problems due to the possible complexity of the environment context where the set of structured, sophisticated, and accurate physics parameters can be hard to acquire.
Furthermore, obtaining the sophisticated physics models and the corresponding systematic physics parameters is challenging and requires expert knowledge \cite{bib:sim-blen-datasets}. To better generalize the dynamics prediction model, Qi et al.~\cite{rpcin} proposed Region Proposal Convolutional Internation Network (RPCIN), a vision-based dynamics prediction model, which simplified the inputs to be a sequence of RGB images and simple object descriptions, e.g., bounding-boxes. RPCIN utilizes convolutional neural network (CNN) and Region of Interest (RoI) Pooling operation \cite{fasterrcnn, maskrcnn} to extract each object state feature within the environment context where an interaction network \cite{interaction_learn_phys} processes all the state features for predicting future state. Despite previous success, such end-to-end vision-based dynamics prediction models, like RPCIN, may find a shortcut to minimize the empirical loss and overfit to the training environment. Thus, the explainability of the model can be poor and the model can suffer from environment misalignment challenges, such as the cross-domain challenge \cite{bib:sim-blen-datasets}.

To address the cross-domain challenge, Xie et al.~\cite{bib:sim-blen-datasets} argued to first map the original visual appearance of both objects of interest and environment context to the abstract space. 
Despite the difference in the appearance details across visual domain, the respected representations in such abstract space stay the same. For example, while the appearance of vehicles can differ between the real world~\cite{citesypies} and video games~\cite{gta}, their semantic segmentation masks, as instances of the abstract space, are the same.
Then, the dynamics prediction is performed on the abstract space so that various visual domains can be aligned. In the scope of the billiard game discussed in \cite{bib:sim-blen-datasets}, the object bounding-box and the semantic segmentation of environment context were used as the instance of abstract space for billiard balls and billiard table, respectively. However, they mainly studied abstracting the environment context, where the discussion on the insights of the usage of bounding-box as the abstract of objects was neglected. We also noticed that, as the empirical studies demonstrated in \cite{bib:sim-blen-datasets}, when replacing the RGB image with the semantic segmentation of the environment context, where the visual information of objects of interest is completely missing, the vision-based model can still maintain an outstanding performance. Under this scenario, since the visual information of object is missing, the bounding-boxes of each object are the only source that contains the object position information. Instead of serving as direct inputs, bounding-box is merely consumed by RoI Pooling operation to fragment the object state features  from the whole outputs of the CNN backbone. Therefore, based on those observations, we hypothesize that the object position information is implicitly \emph{`captured'} by the CNN for dynamics prediction. Therefore, in this paper, we aim to study the insight of using the object bounding-box as the object abstract for dynamics prediction, especially emphasizing the rationale behind the indirect position encoding by performing RoI Pooling according to the object bounding-box. 

Islam et al.~\cite{bib:cnn-position-encoding} discussed that the spatial information is derived from zero padding by utilizing a classification or semantic segmentation pretrained CNN backbone to predict synthesized \emph{gradient-like position map}.
However, their empirical experiments were orthogonal to both classification and semantic segmentation, which are the tasks of the pretrained backbones. Furthermore, they only study the effect of zero padding, where other padding modes are overlooked. Although in a recent work~\cite{Islam2024}, Islam et al. provide additional investigation with more tasks and padding modes, the discussion on the position encoding in the dynamics prediction is still missing.  Nevertheless, those previous works still inspire us to speculate that the position information inherent in object abstract is indirectly \emph{`captured'} by CNN through padding and, then, leveraged for dynamics prediction. Specifically, we hypothesize that the padding enables the CNN to encode position information to its output features. Subsequently, by performing RoI Pooling operation with respect to the the bounding-boxes, the object state features fragmented from the whole CNN output contains object position descriptions. 

\begin{figure*}[!ht]
\centering
     \includegraphics[width=0.85\textwidth]{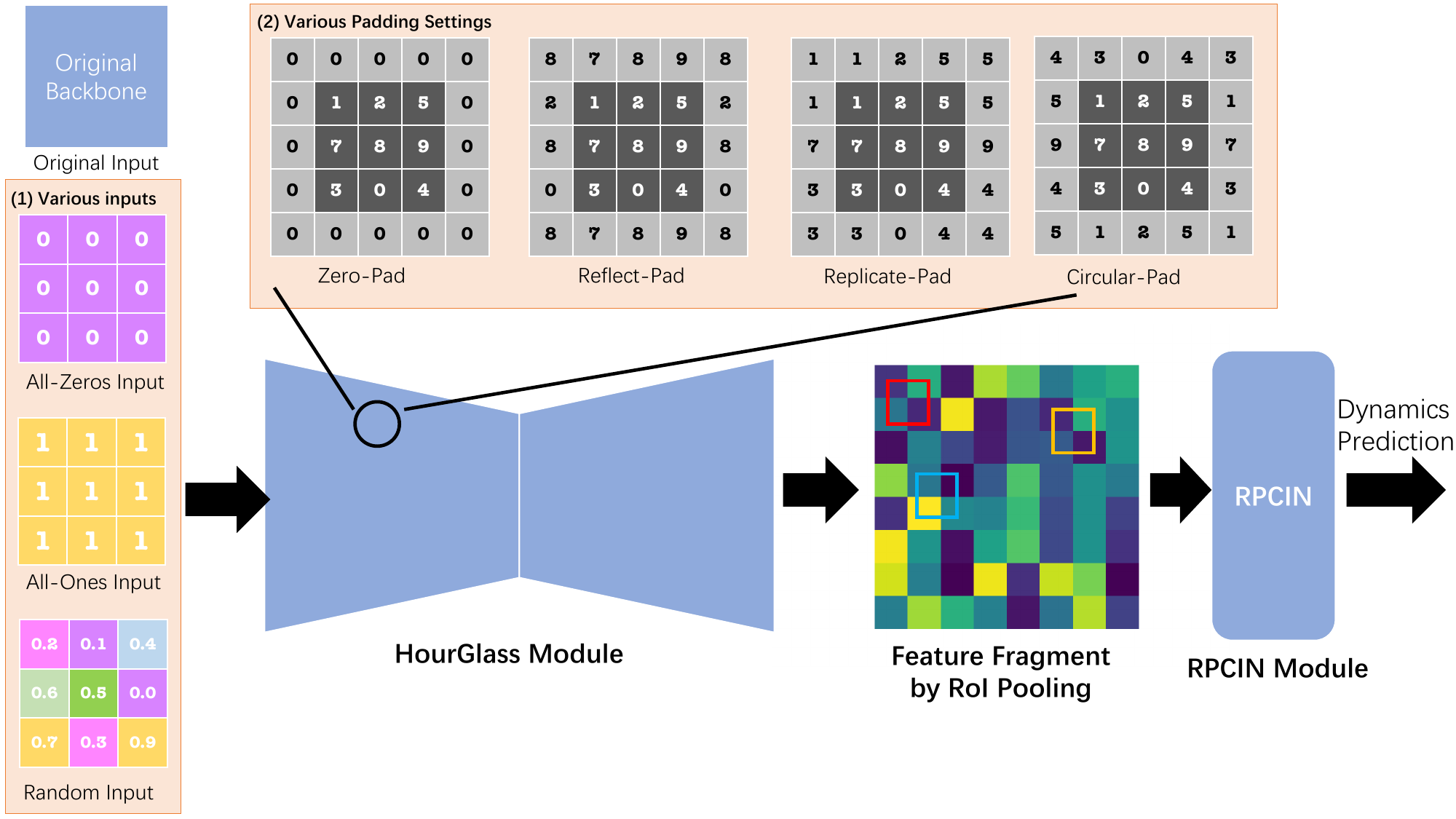}
    \caption{Illustration of our investigation details. 1) We replace the original Hourglass Module input, which is the output of the original backbone, with All-Zeros Input, All-Ones Input, and Random Input to study the effect on the global feature map of various meaningless inputs; 2) We test multiple CNN padding methods in the Hourglass Module, which are Zero-Pad, Reflect-Pad, Replicate-Pad, and Circular-Pad, to study the effect on the the global feature map of various padding setting. Further, we also studied the joint effect of padding modes with or without the bias weights (not illustrated). We include a simple illustration of the Hourglass Module\cite{hourglass} and RPCIN\cite{rpcin} for completion and refer readers to the original papers for detailed illustrations and discussions.} %
    \label{fig:framework}
\end{figure*}

In order to verify our assumption, following \cite{bib:sim-blen-datasets}, we used RPCIN \cite{rpcin} as a probe and conducted experiments on \emph{SimB} dataset proposed by \cite{rpcin} where only dynamics between objects are involved, and all objects share the same physics properties. 
Without loss of generality, despite the simplicity of the dataset, it provides an ideal template for focusing on the discussion of position encoding insight, and the conclusion can also be generalized to a more complex scenario.
To solely utilize bounding-box for providing object position information and thoroughly investigate the process of extracting position information, we replace the visual information with several synthetic inputs, such as all ones, all zeros, or random inputs, and explore the different hyper-parameter settings of padding. Different from classification or semantic segmentation used in \cite{bib:cnn-position-encoding}, position information is critical for a correct dynamics prediction. Therefore, the model performance can directly reveal the capability of the position encoding. Our experiments show that, as also demonstrated in \Cref{fig:illustration},  when the underlying environment context stays unaltered for all scenes, the distinctions between object state features, fragmented from different parts of CNN outputs, are essential for encoding object position information for dynamics prediction. A proper padding setting can lead to such distinctions, and the randomness in the model inputs can also have a similar capability. On the contrary, when the environment context varies, such as \emph{SimB-Border} and \emph{SimB-Split}, merely relying on the position encoding of objects is insufficient for dynamics prediction. By investigating the mechanisms of how models handle different types of input data and environmental contexts, our work sheds light on the adaptability and explainability of AI systems on applications that require position encoding. Furthermore, by understanding the limitations of current approaches in handling various environment contexts, our work also suggests developing an explainable model to encode and process the complex environment context can be a possible future research to improve the performance and generality of dynamics prediction models in real-world applications.

\section{Preliminary}

Following \cite{bib:sim-blen-datasets}, this work focuses on predicting dynamics in a billiard game scenario and using RPCIN \cite{rpcin} as a probe, which is evaluated by both short-term and long-term prediction performance \cite{rpcin}. During the training phase, the model refers to a sequence of $T_{ref}$ consecutive image frames, denoted as ${X_{1-T_{ref}} ... X_{0}}$, along with corresponding reference ball states in the respective frame, denoted as ${S_{1-T_{ref}} ... S_{0}}$. The goal of the model is to predict the ball states for the next $T_{pred}$ frames, represented as ${S'_{1}...S'_{T_{pred}}}$, which are compared with ground-truth states ${S_{1}...S_{T_{pred}}}$ for supervised training. In the inference phase, the model utilizes the sequences ${X_{1-T_{ref}} ... X_{0}}$ and ${S_{1-T_{ref}} ... S_{0}}$ as reference to predict the short-term states ${S'_{1}...S'_{T_{pred}}}$ and the long-term states ${S'_{T_{pred}+1}...S'_{2T_{pred}}}$, where two predictions are evaluated separately. RPCIN \cite{rpcin} was proposed as an end-to-end solution which leverages only the bounding-box information of each ball to represent the frame state $S$. By employing RoI Pooling operation \cite{fastrcnn, maskrcnn}, RPCIN directly fragments and extracts the ball state features $b_i$ from the whole visual features encoded by a CNN backbone from the RGB image for each reference frame. For a comprehensive understanding, we will provide a brief summary of RPCIN, while directing readers to \cite{rpcin} for detailed descriptions and discussions.
As described in \cite{rpcin, bib:sim-blen-datasets}, to infer the dynamics of each ball by utilizing the ball state features $b_i$, RPCIN incorporate Convolutional Interation Network (CIN) which is composed of five CNNs, denoted as $f_O$, $f_R$, $f_A$, $f_Z$, and $f_P$  \cite{rpcin}. 
Firstly, the self-dynamics feature of the $i$-th ball at the $t$-th frame is derived by $f_O$ with $b_i^t$ as input. Correspondingly, the pairwise relative-dynamics feature between $i$-th ball and $j$-th ball in the same frame is computed by $f_R$ with both $b_i^t$ and $b_j^t$ as inputs. Secondly, the overall-dynamics feature $e_i^t$ is derived by $f_A$, where the input is the summation of the self-dynamics feature and all relative-dynamics features with respect to $i$-th ball at $t$-th frame. Thirdly, the static-dynamics features $z_i^t$ is computed by $f_Z$ with $b_i^t$ and $e_i^t$ as inputs. Finally, the state feature of $i$-th ball at the next frame $t+1$ can be predicted by $f_P$ which consumes $z_i$ of previous $T_{ref}$ frames as input. The overall calculation is expressed in \ref{eq:rpcin_model} \cite{rpcin}.
\begin{equation}
\label{eq:rpcin_model}
\begin{aligned}
    e_i^t &= f_A(f_O(b_i^t) + \sum_{j \neq i}f_R(b_i^t, b_j^t)), \\
    z_i^t &= f_Z(b_i^t, e_i^t), \\
    b_i^{t+1} &= f_P(z_i^t, z_i^{t-1},...,z_i^{t-T_{ref}+1})
\end{aligned}
\end{equation}

\section{Investigate How Position Information is Utilized in Dynamics Prediction}

In order to investigate and reveal insight into how the dynamics predictions model solely utilizes bounding-boxes and RoI Pooling operations to indirectly provide spatial information to encode and process position information, we conduct experiments by altering network inputs and modifying padding settings. Specifically, to study the contribution of different network inputs, as shown in \cref{fig:framework}, we replace the meaningful visual features that are extracted from the RGB visual inputs (e.g., video frames) with synthesized features while keeping the environment context consist.
We also modify padding setting to investigate its effect on encoding position information for dynamics prediction.

In the following sections, we will first describe the experiment settings and then discuss how position information is utilized. Furthermore, we provide empirical results to show that when environment context varies, merely relying on object abstracts and the model's capability of indirect position encoding is insufficient for accurate dynamics prediction.

\subsection{Backbone Details Modifications}
\label{sec:detail-modification}
The entire dynamics model is composed of two major components: Hourglass backbone \cite{hourglass} for extracting visual features and CIN module \cite{rpcin} to infer dynamics. Since RoI Pooling operation is applied on the output feature of Hourglass backbone, all spatial information should already be encoded in such output features, where CIN module will simply utilize those features for dynamics prediction. Therefore, our experiments focus on the details and the encoding mechanism within the Hourglass backbone, which lead to the distinctions between output features of different objects that are sufficient for correctly identifying their state.

Hourglass backbone \cite{hourglass} is composed of a CNN with residual module to downsample the RGB input to a smaller scale for reducing computation complexity, and followed by a Hourglass module to refine the visual information \cite{rpcin}. Therefore, to better analyze the influence of network detail on position information encoding and reduce the possible nuisance impact inherent in the network complexity, we focus on the modifications related to the Hourglass module. In detail, our major modifications are made in two folds, as illustrated in \ref{fig:framework}: (1) replacing the meaningful visual inputs to Hourglass module with synthesized inputs, such as \emph{All-Zeros Inputs}, \emph{All-Ones Inputs} and \emph{Random Inputs}, and (2) changing the padding mode, such as \emph{Zero-Pad, Reflect-Pad, Replicated-Pad, and Circular-Pad}, and padding size within Hourglass module. Additionally, to further increase the comprehensiveness of our investigation on the padding mode, we also studied the joint effect of padding modes with and without the bias weights.

\subsection{Datasets and Metrics}
\label{sec:dataset}

\begin{table*}[]
\centering 
\caption{Quantitative comparison of different padding modes \emph{with bias weights} within CNN kernels trained on different types of input. We highlight the performance of the model, which fails to encode position information in \textbf{bold}. P1 and P2 measure the prediction errors for short-term and long-term dynamics prediction, respectively.}
\label{table:with-bias}
\renewcommand{\arraystretch}{1.65}
{
\begin{adjustbox}{width=1.0\textwidth}
\begin{tabular}{c|cc|cc|cc|cc}
\hlineB{3}
Padding Mode (w/ bias) & \multicolumn{2}{c|}{Zero}                 & \multicolumn{2}{c|}{Reflect}                            & \multicolumn{2}{c|}{Replicate}                          & \multicolumn{2}{c}{Circular}                      \\
Eval Period            & P1 $\downarrow$                 & P2 $\downarrow$                   & P1 $\downarrow$                         & P2 $\downarrow$                        & P1 $\downarrow$                         & P2 $\downarrow$                         & P1 $\downarrow$                      & P2 $\downarrow$                      \\ \hline
Visual Inputs          & 2.72\tiny{$\pm$0.31} & 27.94\tiny{$\pm$1.08} & 2.74\tiny{$\pm$0.30}         & 28.43\tiny{$\pm$1.21}        & 2.82\tiny{$\pm$0.42}         & 28.94\tiny{$\pm$1.11}        & 2.73\tiny{$\pm$0.42}         & 28.03\tiny{$\pm$1.09}        \\
All-Zeros Inputs       & 2.97\tiny{$\pm$0.53} & 29.83\tiny{$\pm$1.13}  & \textbf{144.34 \tiny{$\pm$0.21}} & \textbf{145.14\tiny{$\pm$0.31}} & \textbf{144.35\tiny{$\pm$0.31}} & \textbf{145.51\tiny{$\pm$0.30}} & \textbf{144.42\tiny{$\pm$0.30}} & \textbf{145.08\tiny{$\pm$0.31}} \\
All-Ones Inputs        & 3.11\tiny{$\pm$0.49} & 30.48\tiny{$\pm$1.48}  & \textbf{144.43\tiny{$\pm$0.30}} & \textbf{145.17\tiny{$\pm$0.32}} & \textbf{144.43\tiny{$\pm$0.31}} & \textbf{145.17\tiny{$\pm$0.29}} & \textbf{144.43\tiny{$\pm$0.32}} & \textbf{145.09\tiny{$\pm$0.21}} \\
Fixed-Random Inputs    & 2.91\tiny{$\pm$0.47} & 30.03\tiny{$\pm$1.15}  & 3.01\tiny{$\pm$0.42}         & 31.48\tiny{$\pm$1.67}        & 3.04\tiny{$\pm$0.52}         & 29.56\tiny{$\pm$1.08}        & 3.17\tiny{$\pm$0.46}         & 30.46 \tiny{$\pm$1.81}        \\
Random Inputs          & 3.00\tiny{$\pm$0.55} & 29.40\tiny{$\pm$0.96}  & 2.90\tiny{$\pm$0.41}         & 30.68\tiny{$\pm$1.09}      & 3.17\tiny{$\pm$0.48}         & 31.09\tiny{$\pm$1.09}        & 2.98\tiny{$\pm$0.39}         & 29.07\tiny{$\pm$1.73}        \\ \hlineB{3}
\end{tabular}
\end{adjustbox}
}
\end{table*}

To surgically study the process of position information extraction, we conduct experiments on \emph{SimB} dataset \cite{rpcin}, which simulate three balls billiard scenario. There are 1000 video clips for training and testing respectively, where each video clip contains 100 frames. The resolution of each frame is $64 \times 64$. The environment context stays constant for all video frames, all balls have the same physical properties, and the ball objects bounce when hitting image boundaries or other balls. Thus, given the property of the dataset, the object bounding-box, which serves as the object abstract, can provide sufficient position information for accurate dynamics prediction.

In addition, to investigate the deficiency of the object abstract and the limitation of merely relying on the model's mechanism of indirect position encoding, we evaluate the model performance when only object abstract is utilized on datasets extended from \emph{SimB} proposed by \cite{bib:sim-blen-datasets}: \emph{SimB-Border} and \emph{SimB-Split}. \emph{SimB-Border} increases the image resolution to $192 \times 96$ and adds borders to the image boundaries, where the size of borders is randomly selected as integers in the range of $[0, 15]$ and is fixed for all frames in one video. To further increase the prediction difficulty, \emph{SimB-Split} adds five-pixels wide vertical bar into the scene of \emph{SimB-Border}, where the center of the vertical bar is placed at a location randomly chosen as an integer in the range of $[64, 128]$ and kept constant over all frames in one video. To train the model, following \cite{rpcin,bib:sim-blen-datasets}, the length of the reference frame is set to four, and the length of training prediction frames is set to 20. For evaluation, performances of short term predictions $\{1,...T_{pred}\}$ (P1) and long term predictions $\{T_{pred+1},...2T_{pred}\}$ (P2) are separately evaluated, where squared $l_{2}$ distance between predictions and ground-truth are scaled by $1000$ to be used as evaluation metric. Hyper-parameters settings, other than the studies we focus on, stay the same with  \cite{rpcin,bib:sim-blen-datasets}.

\subsection{How Position Information Is Utilized}
\label{sec:position-information}
 As discussed in \Cref{sec:detail-modification}, in order to remove environment visual information and narrow the model focus on object abstracts, we replace the meaningful visual features, \emph{Visual Inputs}, with four types of synthesized inputs:  \emph{All-Zeros Inputs}, \emph{All-Ones Inputs}, \emph{Fixed-Random Inputs} and \emph{Random Inputs}. \emph{All-Zeros Inputs} and \emph{All-Ones Inputs} are features of all values of zero or one with the same size of \emph{Visual Inputs}. \emph{Fixed-Random Inputs} and \emph{Random Inputs} are both features with randomly generated values with the same size as \emph{Visual Inputs}. \emph{Fixed-Random Inputs} only generate such random features once before training and stay the same throughout the training process, whereas \emph{Random Inputs} randomly generate such features for each iteration.
 For the selections of padding mode, we conduct experiments with four padding modes provided by PyTorch \cite{pytorch}: \emph{Zero Pad}, \emph{Reflect Pad}, \emph{Replicate Pad} and \emph{Circular Pad}. 
 Furthermore, we investigate the combined effect of using different padding modes with or without the bias weights in the CNN kernels.

\begin{table*}[]
\centering
\caption{Quantitative comparison of different padding modes \emph{without bias weights} within CNN kernels trained on different types of input. We highlight the performance of the model, which fails to encode position information in \textbf{bold}. P1 and P2 measure the prediction errors for short-term and long-term dynamics prediction, respectively.}
\label{table:without-bias}
\renewcommand{\arraystretch}{1.65}
{
\begin{adjustbox}{width=1.0\textwidth}

\begin{tabular}{c|cc|cc|cc|cc}
\hlineB{3}
Padding Mode (w/o bias) & \multicolumn{2}{c|}{Zero}                           & \multicolumn{2}{c|}{Reflect}                        & \multicolumn{2}{c|}{Replicate}                      & \multicolumn{2}{c}{Circular}                        \\
Eval Period             & P1 $\downarrow$                      & P2 $\downarrow$                       & P1 $\downarrow$                       & P2 $\downarrow$                       & P1 $\downarrow$                       & P2 $\downarrow$                       & P1 $\downarrow$                       & P2 $\downarrow$                       \\ \hline
Visual Inputs           & 2.82\tiny{$\pm$0.36}         & 28.31\tiny{$\pm$1.31}        & 2.84\tiny{$\pm$0.33}         & 29.02\tiny{$\pm$0.91}        & 2.88\tiny{$\pm$0.34}         & 29.02\tiny{$\pm$1.01}        & 2.93\tiny{$\pm$0.63}         & 29.95\tiny{$\pm$1.46}        \\
All-Zero Inputs         & \textbf{144.41\tiny{$\pm$0.21}} & \textbf{145.13\tiny{$\pm$0.11}} & \textbf{144.43\tiny{$\pm$0.29}} & \textbf{145.27\tiny{$\pm$0.34}} & \textbf{144.31\tiny{$\pm$0.20}} & \textbf{145.14\tiny{$\pm$0.27}} & \textbf{144.42\tiny{$\pm$0.31}} & \textbf{145.07\tiny{$\pm$0.29}} \\
All-Ones Inputs         & 3.36\tiny{$\pm$0.45}         & 31.45\tiny{$\pm$1.37}        & \textbf{144.37\tiny{$\pm$0.21}} & \textbf{145.17\tiny{$\pm$0.37}} & \textbf{144.43\tiny{$\pm$0.29}} & \textbf{145.08\tiny{$\pm$0.21}} & \textbf{144.43\tiny{$\pm$0.36}} & \textbf{145.14\tiny{$\pm$0.30}} \\
Fixed-Random Inputs     & 3.15\tiny{$\pm$0.39}         & 31.83\tiny{$\pm$0.96}        & 3.26\tiny{$\pm$0.40}         & 31.98\tiny{$\pm$1.14}        & 3.22\tiny{$\pm$0.52}         & 30.56\tiny{$\pm$1.51}        & 3.21\tiny{$\pm$0.47}         & 31.46\tiny{$\pm$1.31}        \\
Random inputs           & 3.19\tiny{$\pm$0.42}         & 31.36\tiny{$\pm$1.10}        & 3.09\tiny{$\pm$0.32}         & 31.03\tiny{$\pm$1.25}        & 3.08\tiny{$\pm$0.44}         & 31.12\tiny{$\pm$1.46}        & 3.02\tiny{$\pm$0.35}         & 30.07\tiny{$\pm$0.58}        \\ \hlineB{3}
\end{tabular}
\end{adjustbox}
}
\end{table*}

As shown in \Cref{table:with-bias}, when including the bias weights within CNN kernels and utilizing default padding mode (\emph{Zero Pad}) \cite{rpcin}, compared to \emph{Visual Inputs}, all synthesized inputs achieve comparable performance on both short-term and long-term predictions. This implies that, even without visual information, the object abstracts can provide sufficient position information for dynamics prediction.  By further examining the \Cref{table:with-bias} and checking the performance of different combinations of padding modes and inputs, all combinations with \emph{Random Inputs} can achieve good performance whereas the constant inputs can only work with \emph{Zero Pad} and fail on all other padding modes. 

In order to better understand the insights behind the numerical results, we visualize the sample output features, as demonstrated in \cref{fig:illustration}. The visualizations suggest that creating an inconsistency across the output feature space of the Hourglass module is necessary for fragmenting object state features with distinct values corresponding to bounding-boxes at different locations. When the environment context stays constant, the state features represented by possibly random but distinct numerical values enable the model to implicitly encode sufficient position information for accurate dynamics prediction.
Since \emph{Random inputs} already create such inconsistency in the input space, models with various padding modes can all satisfy such necessity. Contrarily, when inputs are constant and bias weights is utilized in CNN kernels, only \emph{Zero Pad} can achieve good performance, and other padding modes fail to create such inconsistency across the output feature space of Hourglass module. Since the failed padding modes simply copy the value of the edge features, the features after padding are still the same everywhere. Noticeably, when inputs are \emph{All-Zeros}, the model with \emph{Zero Pad} can still achieve such inconsistency. This is due to the fact that by using bias weight within the CNN kernel, the output value of the first CNN layer will not be all zeros. Thus, inconsistency will be created by the  \emph{Zero Pad} in the following CNN layers, which allows the model to achieve good performance. As validated in \Cref{table:without-bias},
removing bias weights from CNN kernels results in unsatisfactory performance for the combination of \emph{Zero Pad} and \emph{All-Zeros Inputs}, as well as for combinations of other padding modes and constant inputs.
\begin{figure*}[!ht]
\centering
     \includegraphics[width=0.85\textwidth]{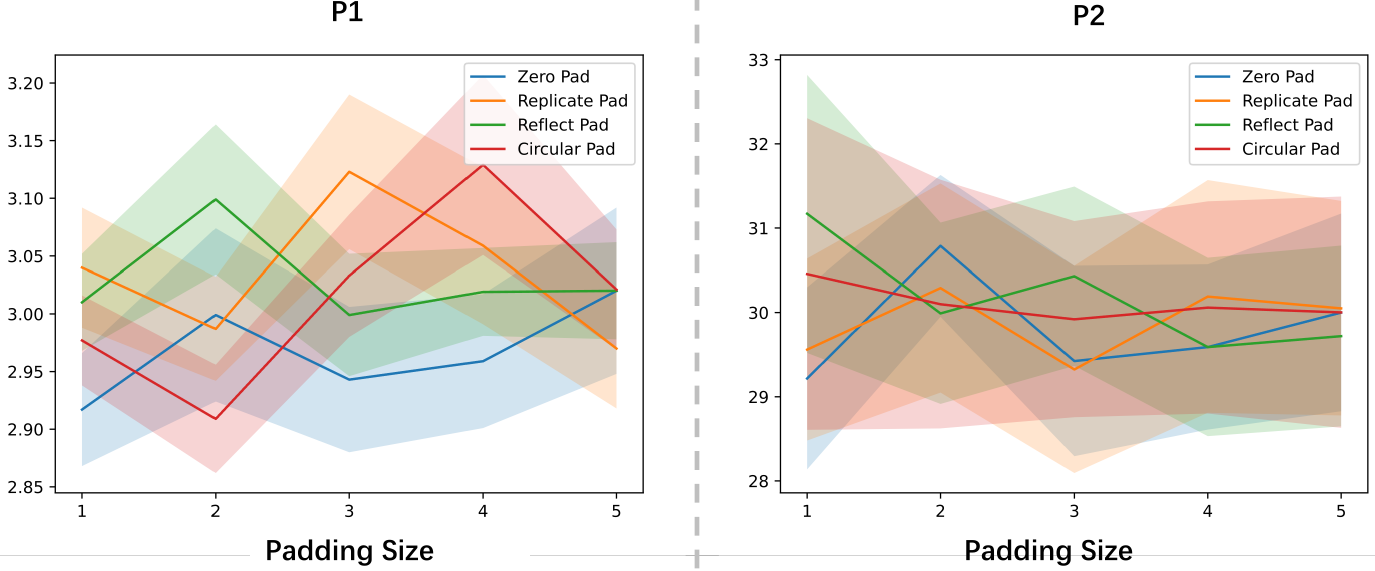}
    \caption{Quantitative comparison between different padding modes and padding size with bias weight trained on \emph{Fixed-Random Inputs}. We repeat the experiments of each padding mode with 10 trials. This quantitative results reveal that when bias weight is incorporated, models with different padding modes show comparable performance. P1 and P2 measure the prediction errors for short-term and long-term dynamics prediction, respectively.} %
    \label{fig:padding-size}
\end{figure*}

As previously discussed, \emph{Random Inputs} alone can provide sufficient inconsistency so that good dynamics prediction performance can be yielded with all padding modes even without the bias weights in the CNN kernels.

\subsection{How Padding Hyper-Parameters Affect the Encoding}
As discussed in \Cref{sec:position-information}, when object abstracts are solely available, the position information can be inferred if inconsistency across output features of the Hourglass module exists. To further investigate the ramifications of changing hyper-parameters of padding for dynamics prediction, we conduct comprehensive experiments by both altering the mode of the padding and changing the size of the padding. Following experiments discussed in \Cref{sec:position-information}, to empower position information encoding for all padding modes while removing visual information, we replace the \emph{Visual Inputs} with \emph{Fixed-Random Inputs}.  As shown in \Cref{fig:padding-size}, changing the padding size will not significantly affect the dynamics prediction performance. This implies that as long as aforementioned inconsistency exists on the output features space of the Hourglass module, sufficient position information can be encoded in object state features for correct dynamics prediction.

\subsection{When Environment Information Is Necessary}
Our previous experiments empirically demonstrate that position information can be inferred from object abstracts. Such position information is provided by inconsistencies across the output feature space of the Hourglass module, which can be created by either proper padding settings or inconsistencies in the inputs.
However, as discussed in \cite{bib:sim-blen-datasets}, in order to accurately predict object dynamics on \emph{SimB-Border} and \emph{SimB-Split}, methods are required to utilize the environment context information. Therefore, solely relying on object abstracts will not be sufficient for accurate dynamics prediction. To verify such insufficiency, we replace \emph{Visual Inputs} with various synthesized inputs and evaluate different padding modes, similar to previous studies in this paper. The results are shown in \Cref{table:context-input,table:context-pad}, respectively.

 \begin{table*}[t]
\centering 
\caption{Quantitative comparison of various inputs on \emph{SimB-Border} and \emph{SimB-Split} datasets. Results reported in \cite{bib:sim-blen-datasets} was used as the \emph{Visual Inputs} Performance. The best results are highlighted in \textbf{bold}. P1 and P2 measure the prediction errors for short-term and long-term dynamics prediction, respectively.}\label{table:context-input}
\renewcommand{\arraystretch}{1.5}
{
\begin{adjustbox}{width=0.65\linewidth}
\begin{tabular}{ccc|cc}
\hlineB{3}
Dataset           & \multicolumn{2}{c|}{SimB-Border} & \multicolumn{2}{c}{SimB-Split} \\ \cline{2-5} 
Eval Period       & P1 $\downarrow$             & P2 $\downarrow$              & P1 $\downarrow$            & P2 $\downarrow$             \\ \hline
Visual Inputs~\cite{bib:sim-blen-datasets}     & \textbf{1.13$\pm$0.01}   & \textbf{9.57$\pm$0.12}    & \textbf{0.91$\pm$0.02}  & \textbf{7.73$\pm$0.21}   \\
All-Zero Inputs   & 2.04$\pm$0.02   & 11.89$\pm$0.22   & 3.68$\pm$0.05  & 16.85$\pm$0.13  \\
Fixed-Random Inputs & 2.05$\pm$0.02   & 12.58$\pm$0.20   & 3.65$\pm$0.03  & 16.86$\pm$0.06  \\ \hlineB{3}
\end{tabular}
\end{adjustbox}
}
\end{table*}

\begin{table*}[t]
\centering 
\caption{Quantitative comparison of different padding modes with \emph{Fix-Random Inputs} on \emph{SimB-Border} and \emph{SimB-Split} datasets. \emph{Baseline} is the \emph{Visual Input} with \emph{Zero Pad} mode, and we use the performance reported in \cite{bib:sim-blen-datasets}. The best results are highlighted in \textbf{bold}. P1 and P2 measure the prediction errors for short-term and long-term dynamics prediction, respectively. }\label{table:context-pad}
\renewcommand{\arraystretch}{1.4}
{
\begin{adjustbox}{width=0.6\linewidth}
\begin{tabular}{ccc|cc}
\hlineB{3}
Dataset     & \multicolumn{2}{c|}{SimB-Border} & \multicolumn{2}{c}{SimB-Split} \\ \cline{2-5} 
Eval Period & P1 $\downarrow$             & P2 $\downarrow$              & P1 $\downarrow$            & P2 $\downarrow$             \\ \hline
Baseline~\cite{bib:sim-blen-datasets}    & \textbf{1.13$\pm$0.01}     & \textbf{9.57$\pm$0.12}      & \textbf{0.91$\pm$0.02}    & \textbf{7.73$\pm$0.21}     \\
Zero        & 2.05$\pm$0.02     & 12.58$\pm$0.20     & 3.65$\pm$0.03    & 16.86$\pm$0.06    \\
Reflect     & 2.04$\pm$0.02     & 12.23$\pm$0.25     & 3.61$\pm$0.05    & 16.71$\pm$0.38    \\
Replicate   & 2.04$\pm$0.01     & 11.93$\pm$0.26     & 3.61$\pm$0.05    & 16.96$\pm$0.68    \\
Circular    & 2.04$\pm$0.01     & 12.34$\pm$0.08     & 3.63$\pm$0.04    & 16.55$\pm$0.47    \\ \hlineB{3}
\end{tabular}
\end{adjustbox}
}
\end{table*}

When considering the environment context is critical, such as the environment context varies between videos, models merely utilizing object abstracts only achieve mediocre performance. Furthermore, as shown in \Cref{table:context-input,table:context-pad}, the model performance becomes worse when the environment context becomes more complex, i.e., the evaluation on \emph{SimB-Split} where there is a vertical splitting bar randomly located in the scene.

\subsection{Discussion Summary and Broader Impact}
In this work, we demonstrate and analyze that when bounding box and region of interest pooling are used to indirectly provide location information via feature fragmenting, the distinctions between each object feature fragment are necessary to encode object position information. Such distinctions can be created by proper CNN padding or inconsistency in the input to the network backbone, as shown in \Cref{table:without-bias,table:with-bias,fig:illustration}. Further, we emphasize that despite CNN's ability to implicitly encode position information, solely relying on the feature fragments distinction for dynamics prediction, without utilizing any visual information, may only be valid when the environment context stays the same. Should the environment context change, visual input containing environment context is necessary for the vision-based dynamics prediction models to encode sufficient information beyond the position of the object of interest for accurate dynamics prediction. Aside from our discussions on SimB, we see that the discoveries of our work have the potential to be generalized to other visual domains, e.g., real-world domain because our study does not focus on any characteristic of a specific visual domain. 
The insights provided by this study on how neural networks encode and utilize position information, even in an indirect manner, have broader implications for the explainability and generalizability of AI systems. Since the research community seeks to develop more versatile and adaptable DNNs, understanding the mechanisms by which they represent and process fundamental information like spatial relationships is crucial. This knowledge can inform the design of more robust and flexible architectures capable of handling a wider range of tasks and environments, thus contributing to the overall scalability and generalizability of AI systems.
Therefore, we believe that beyond the scope of dynamics prediction, our work can also benefit other research fields where correctly encoding position information is essential, such as autonomous driving \cite{DBLP:journals/corr/abs-1906-05113}, causal inference with position information \cite{zhu2024diffusioncounterfactualsinferringhighdimensionalcounterfactuals} and visual question answering \cite{WU201721}. 

Furthermore, in this work, we seek to analyze the position encoding mechanism by utilizing a simple but controlled dataset and surgically modify the model backbone to focus our discussion. The primary empirical results shown in \cref{table:with-bias,table:without-bias} and the visualization in \cref{fig:illustration} not only reveal the mechanisms of position encoding in dynamics prediction models but also contribute to the broader goal of analyzing the explainability of DNNs. By revealing how different input types and network configurations affect the model's ability to encode spatial information under a controlled environment, we gain insights into the internal representations formed by these networks. This approach demonstrates how targeted modifications and analyses can unpack the complex information processing occurring within DNNs, and we hope our work can encourage future explainable AI researchers to also conduct simple, controllable, and targeted analysis, in addition to the studies on the complex dataset that involves many sophisticated uncertainties.

\section{Conclusion}
In this work, utilizing RPCIN and billiard games as a probe, we comprehensively investigate the process of position information encoding for vision-based dynamics prediction, where only object abstracts, i.e., bounding-boxes, are indirectly utilized while the environment stays unchanged. The empirical results reveal that the inconsistency in the distribution of output features is the key to empowering the model to encode the position information. Such inconsistency can be brought up by either the proper padding setting within CNN kernels or the divergence that existed in the original inputs. In addition, our experiments further show that when the environment context varies, merely incorporating object abstracts is insufficient for correct dynamics prediction, where the model performance is jeopardized when the environment context becomes complex. The findings of this study not only contribute to the field of vision-based dynamics prediction but also offer insights into the explainability of AI systems. By elucidating how neural networks encode and utilize position information, this work contributes to the foundation for developing more adaptable and versatile DNNs. The observed limitations in handling varying environment contexts highlight areas where future research could focus on enhancing the robustness and generalizability of models, and ultimately expanding their applicability across diverse domains and tasks.

\bibliographystyle{splncs04}
\bibliography{main}
\end{document}